\pdfoutput=1
\documentclass[runningheads]{llncs}
\usepackage[T1]{fontenc}
\usepackage{graphicx}
\usepackage{ragged2e} 
\usepackage{booktabs,makecell, multirow, tabularx}
\usepackage{subfigure}
\usepackage{amsmath}
\usepackage{float}

\usepackage{geometry}
\begin{document}
%
\title{End-to-end Remote Sensing Change Detection of Unregistered Bi-temporal Images for Natural Disasters}
\titlerunning{End-to-end Change Detection of Unregistered Bi-temporal Images}
%
\author{Guiqin Zhao \and
Lianlei Shan \and
Weiqiang Wang \thanks{Corresponding author}}
\authorrunning{G. Zhao et al.}
%
\institute{School of Computer Science and Technology, University of Chinese Academy of Sciences, Beijing 100049, China\\ \email{{zhaoguiqin20,shanlianlei18}@mails.ucas.edu.cn,wqwang@ucas.ac.cn}\\
}
\maketitle              
\begin{abstract}

Change detection based on remote sensing images has been a prominent area of interest in the field of remote sensing. Deep networks have demonstrated significant success in detecting changes in bi-temporal remote sensing images and have found applications in various fields. Given the degradation of natural environments and the frequent occurrence of natural disasters, accurately and swiftly identifying damaged buildings in disaster-stricken areas through remote sensing images holds immense significance. This paper aims to investigate change detection specifically for natural disasters.
Considering that existing public datasets used in change detection research are registered, which does not align with the practical scenario where bi-temporal images are not matched, this paper introduces an unregistered end-to-end change detection synthetic dataset called xBD-E2ECD. Furthermore, we propose an end-to-end change detection network named E2ECDNet, which takes an unregistered bi-temporal image pair as input and simultaneously generates the flow field prediction result and the change detection prediction result. It is worth noting that our E2ECDNet also supports change detection for registered image pairs, as registration can be seen as a special case of non-registration. Additionally, this paper redefines the criteria for correctly predicting a positive case and introduces neighborhood-based change detection evaluation metrics. The experimental results have demonstrated significant improvements.

\keywords{Change detection \and Registration \and End-to-end \and Synthetic dataset \and Remote sensing.}
\end{abstract}

\section{Introduction}

The purpose of change detection in remote sensing images is to identify differences in objects or natural phenomena between different states. This is achieved by comparing images taken at different times but from the same geographic area, resulting in the generation of a change map. The research on change detection based on remote sensing images has been a prominent and long-standing focus in the field. It plays a crucial role in providing valuable information for decision-making in various domains, such as urban expansion \cite{6948368}, regional changes in lake surfaces \cite{rs6054173}, and damage assessment \cite{2019BuildingDamage}.

In light of the degradation of the natural environment and the frequent occurrence of natural disasters, accurately and swiftly identifying damaged buildings in disaster areas through remote sensing images holds immense significance for analyzing the disaster situation and guiding subsequent emergency rescue efforts. However, existing change detection datasets are pre-registered using tools, which does not align with the practical scenario where images obtained at different times may not be well-matched due to variations in azimuth and shooting angles. In the event of a natural disaster, immediate change detection is desired, but separate registration would consume additional time. Furthermore, registration using tools requires specific information such as geographic coordinates and projection details, which may not always be available or may have been lost.
Therefore, this paper aims to investigate end-to-end change detection for natural disasters, where bi-temporal image registration and change detection are simultaneously performed using deep networks. To the best of our knowledge, we are the first to focus on change detection of unregistered image pairs. It is important to note that registration can be considered a special case of non-registration, and our approach is also applicable to change detection for registered image pairs.

As existing public datasets for change detection are all registered, collecting unregistered bi-temporal image pairs and manually annotating them would require substantial time and effort. To address this, this paper introduces the xBD-E2ECD dataset, a synthetic dataset specifically designed for unregistered end-to-end change detection in natural disasters. Additionally, the paper proposes the E2ECDNet, an end-to-end change detection network that simultaneously predicts the flow field and change detection for unregistered bi-temporal image pairs.
In contrast to single registration and single change detection approaches, the end-to-end framework in this paper considers the interplay between change detection and registration results. The presence of changing pixels increases the complexity of image registration, while the inaccuracies in image registration further complicate change detection. As image registration cannot achieve perfect accuracy, it is highly likely that pixels in the vicinity of a positive pixel are predicted as belonging to the changed class. To address this, the paper redefines the case of correctly predicting a positive and proposes neighborhood-based change detection evaluation criteria. The experimental results demonstrate significant improvements.
 
Deep networks have achieved significant success in both change detection and image registration. However, to the best of our knowledge, there is currently no research on integrating registration and change detection. Therefore, the objective of this paper is to investigate the end-to-end change detection of unregistered image pairs for natural disasters. The main contributions of this study are as follows:
\begin{itemize} 
    \item[$\cdot$] We construct the xBD-E2ECD dataset, a synthetic dataset specifically designed for unregistered end-to-end change detection in natural disasters. This dataset is built upon the xBD dataset \cite{Gupta2019CreatingXA}.  
    \item[$\cdot$] We propose the E2ECDNet, an end-to-end change detection network that takes unregistered bi-temporal image pairs as input and simultaneously predicts the flow field and change detection.
    \item[$\cdot$] We redefine the criteria for correctly predicting a positive case and introduce neighborhood-based change detection evaluation criteria specifically tailored for end-to-end change detection.
\end{itemize}

\section{Related Work}
\subsection{Image Registration} 
Image registration aims to establish pixel-to-pixel correspondence between two images and warp the floating image to align with the reference image based on their spatial mapping. Convolutional neural networks (CNNs) have been widely used to solve the problem of image correspondence. Given an input image pair, CNNs can generate pixel-level dense correspondence known as the flow field\cite{zhao2022explore,zhao2023flowtext,zhao2023generative}. By rearranging the pixels of the floating image according to the flow field, it can be aligned with the reference image.
Dosovitskiy et al. \cite{7410673} constructed the first trainable CNN for light flow estimation, called FlowNet, which utilized a U-Net denoising autoencoder architecture \cite{10.1145/1390156.1390294}. PWC-Net \cite{8579029,8621052}, LiteFlowNet \cite{8579034}, and LiteFlowNet2 \cite{hui20liteflownet2} employed multiple constrained correlation layers on a feature pyramid. These networks warp features at each level using the current flow estimate, resulting in a more compact and efficient architecture. GLUNet \cite{2020GLU} combines global and local correlations to achieve more accurate dense correspondence prediction. Rocco et al. \cite{9167479} improved the performance of the global correlation layer by proposing an end-to-end training neighbor consensus network, NC-Net, which filters out ambiguous matches and preserves locally and cyclically consistent matches.

\subsection{Change Detection} 
Change detection aims to predict whether there are pixel-level changes between the input image pair. Deep networks have achieved remarkable success in bi-temporal remote sensing image change detection. Zhan et al. \cite{8022932} were the first to introduce the siamese convolutional network to address the problem of bi-temporal remote sensing image change detection, achieving excellent results. Since then, several change detection methods based on siamese structures have been proposed, yielding good detection outcomes. Daudt et al. \cite{2018Fully} designed three effective change detection architectures based on fully convolutional neural networks (FCNN): Fully Convolutional Early Fusion (FC-EF), Fully Convolutional Siamese-Concatenation (FC-Siam-conc), and Fully Convolutional Siamese-Difference (FC-Siam-diff). The main difference among these networks lies in the way they fuse bi-temporal information.
Subsequently, to capture global information in space-time, some methods have addressed the limitations of convolutional operations by enlarging the receptive field (RF). For instance, \cite{2020A,2020DASNet} employ ResNet \cite{7780459} as the backbone network, \cite{2019Triplet} replaces traditional convolution with dilated convolution, and \cite{9355573,9254128} incorporate attention mechanisms. Moreover, BiT \cite{9491802} and changeFormer \cite{Bandara2022} further introduce the transformer to enhance change detection performance.

\section{Approach}
\subsection{End-to-End Change Detection Dataset:xBD-E2ECD}
In this paper, we propose an end-to-end change detection dataset called xBD-E2ECD, which is synthesized using random affine transformations based on xBD \cite{9}, a dataset for natural disaster building detection. xBD is a large-scale dataset designed for building damage assessment and covers ten natural disaster events, including hurricane-florence, hurricane-matthew, mexico-earthquake, midwest-flooding, socal-fire, santa-rosa-wildfire, hurricane-michael, hurricane harvey, palu-tsunami, and guatemala volcano. It provides pre-event and post-event satellite images with a size of 1024 × 1024, and each image contains at least one building. The dataset includes four damage types: no damage, minor damage, serious damage, and destructive damage. For post-event images, both the vertex coordinates and the damage type for each building polygon are recorded. For pre-event images, only the vertex coordinates for each building polygon are recorded, indicating that the buildings in the pre-event images are assumed to have no damage. 

\subsubsection{Algorithm for Establishing xBD-E2ECD.}
The pre-event and post-event images in xBD \cite{Gupta2019CreatingXA} are already well registered, but xBD only provides annotations for each building in each individual pre-event or post-event image, without directly providing change detection annotations for each pre-event and post-event image pair. Therefore, the algorithm for synthesizing xBD-E2ECD in this paper is carried out in two steps.
First, we establish the change detection dataset xBD-CD by generating change detection labels for each registered pre-event and post-event image pair. Specifically, we consider the building areas that do not overlap before and after the event, as well as the building areas that overlap but have different damage types before and after the event, as the change areas. 
Second, based on xBD-CD, we synthesize the unregistered end-to-end change detection dataset xBD-E2ECD using random affine transformations. These transformations simulate the differences in altitude and azimuth of the aircraft when capturing images at two different times. Additionally, we record the coordinate offsets after the transformation as flow field annotations and generate masks that mark valid positions as one and invalid positions as zero. A position in the transformed image is considered valid if it corresponds to a position in the image before the transformation. The training set, test set, and hold set are divided according to the original split of xBD \cite{Gupta2019CreatingXA}. 

Furthermore, we analyze the frequency distribution of images with different numbers of positive pixels for each disaster type using histograms. We observe that there is a significant number of images with fewer than 10 or 100 positive pixels. To alleviate data imbalance, we remove the image pairs with fewer than 100 positive pixels and obtain the final dataset. The final distribution is shown in Table \ref{xBD-E2ECD-after-clean}.

\begin{table}[!ht]
\caption{Data distribution under different disaster types. I(image) represents the number of images, P(positive) represents the number of pixels belonging to the changed class in the valid area and N(negative) represents the number of pixels belonging to the unchanged class in the valid area.}
\label{xBD-E2ECD-after-clean}
\centering
\begin{tabular}{ c|l|l|l|l|l }
 \hline
 event & -- & train & hold & test & total \\
 \hline
\multirow{3}{*}{\shortstack{hurricane-\\florence}} & I & 579& 189	& 183 & 951 \\
& P & 1218960 &	297029	& 247438 & 1763427 \\
& N	& 31161032 & 10355293 & 9934175	& 51450500 \\
\hline
\multirow{3}{*}{\shortstack{hurricane-\\matthew}} & I & 1389 & 550 & 417 & 2356 \\
& P & 3146652 & 1591310	& 1029829 & 5767791 \\
& N	& 74478688 & 28932865 & 22343437 & 125754990 \\
\hline
\multirow{3}{*}{\shortstack{mexico-\\earthquake}} &I & 74 & 17 & 50 & 141\\
& P & 153040 & 45563 & 97266 & 295869\\
& N	& 3959912 &	879175 & 2707642 & 7546729 \\
\hline
\multirow{3}{*}{\shortstack{midwest-\\flooding}} &I & 187	& 61 & 68 & 316\\
& P & 243350 & 107143 & 120906 & 471399\\
& N	& 10274063 & 3384751 & 3682900 & 17341714 \\
\hline
\multirow{3}{*}{\shortstack{socal-\\fire}} &I & 532 & 180 & 199 & 911\\
& P & 3477243 & 1346334	& 999303 & 5822880\\
& N	& 29026719 & 9845793 & 10960984	& 49833496 \\
\hline
\multirow{3}{*}{\shortstack{santa-rosa-\\wildfire}} &I & 994 &	352	& 250 & 1596\\
& P & 803318 & 308667 & 335481 & 1447466\\
& N	& 51830442	& 18219140 & 12824843 & 82874425 \\
\hline
\multirow{3}{*}{\shortstack{hurricane-\\michael}} &I & 2739 & 868 & 753 &  4360\\
& P & 8820227 & 2824703	& 2272364 & 13917294\\
& N	& 143622411	& 45685385	& 39663651	& 228971447 \\
\hline
\multirow{3}{*}{\shortstack{hurricane-\\harvey}} &I & 1598	& 544 & 490	& 2632\\
& P & 13093228	& 4427544 & 3875806 & 21396578\\
& N	& 73694683 & 25385895 & 23031376 & 122111954 \\
\hline
\multirow{3}{*}{\shortstack{palu-\\tsunami}} &I & 378 & 132 & 130 & 640\\
& P & 2142905 & 613656 & 726770 & 3483331\\
& N	& 18601299 & 6676299 & 6560098 & 31837696 \\
\hline
\multirow{3}{*}{\shortstack{guatemala-\\volcano}} &I & 10 & 12 & 3 & 25\\
& P & 12538	& 36275	& 2904	& 51717\\
& N	& 551573 & 672670 & 168985 & 1393228 \\
\hline
\multirow{3}{*}{all} &I & 8480 & 2905	& 2543 & 13928\\
& P & 33111461 & 11598224 & 9708067	& 54417752\\
& N	& 437200822	& 150037266	& 131878091	& 719116179 \\
\hline
\end{tabular}
\end{table}

\subsection{End-to-End Change Detection Network: E2ECDNet}
\textbf{Problem Definition}
Let $I_s\in R^{H\times W \times3}$ and $I_t\in R^{H\times W \times3}$ denote the pre-event image (source image) and the post-event image (target image), respectively. For an unregistered input image pair $(I_s,I_t)$, the goal is to output the flow field $w\in R^{H\times W \times2}$ for image registration and the change detection probability map $p\in R^{H\times W \times2}$ for change detection.
The flow field $w\in R^{H\times W\times 2}$ represents the pixel-level two-dimensional motion vector, which allows us to warp the source image to the target image using the following equation:
\begin{equation}
I_{t}(x) \approx I_{s}(x + w(x)).
\end{equation}
The change detection probability map $p\in R^{H\times W\times 2}$ indicates the probability of each pixel in the target image being classified as changed or unchanged.

\begin{figure}[htbp]
\centering
\includegraphics[width=0.8\textwidth]{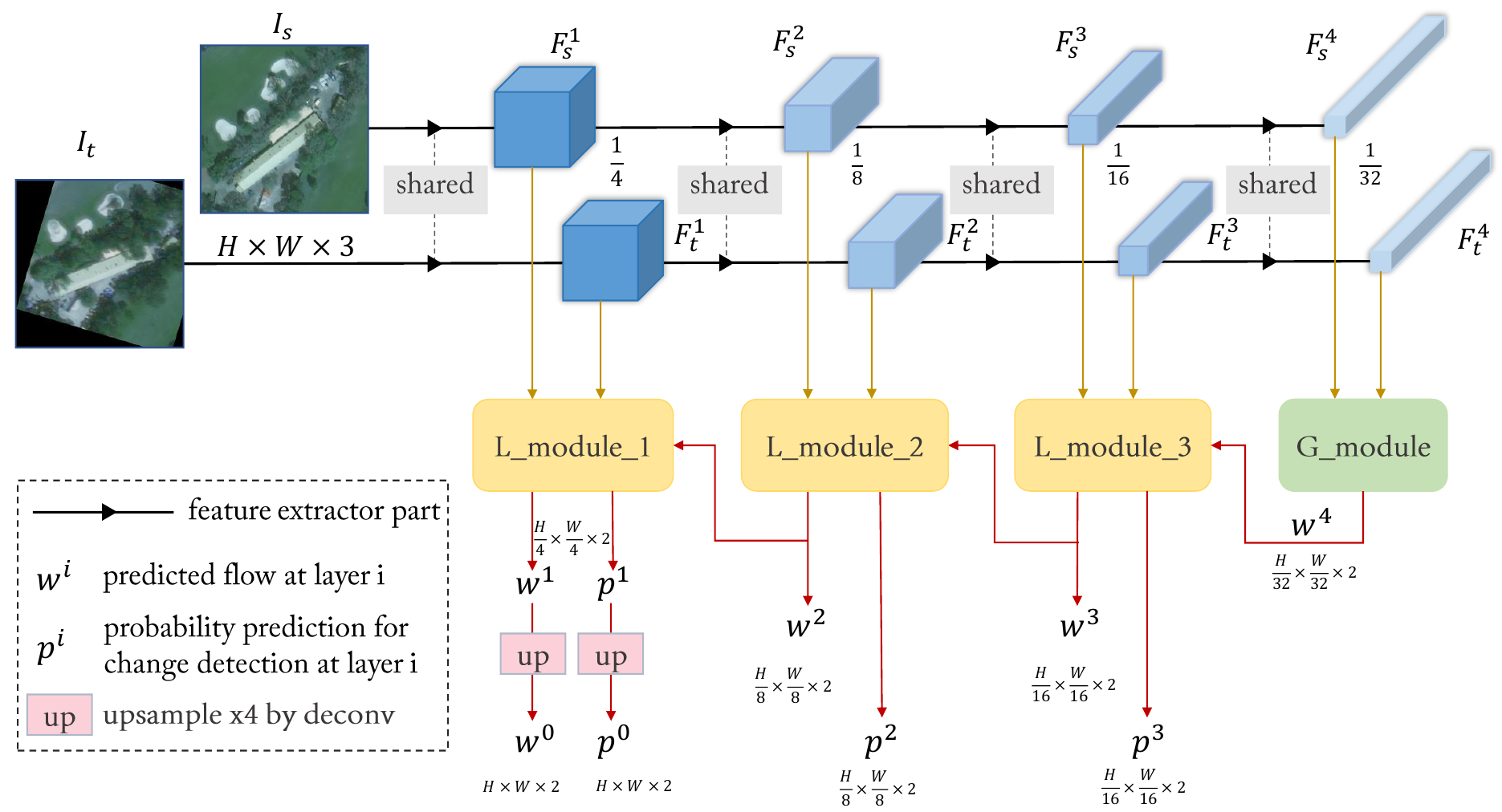}
\caption{The overall structure of the proposed net E2ECDNet. Multiscale feature maps are used to predict the flow field and change detection map.}
\label{overview}
\end{figure}

\textbf{Model Architecture}
The overall structure of our model is illustrated in Figure \ref{overview}. Firstly, the source image $I_s$ and the target image $I_t$ are separately input into the backbone network with shared weights, resulting in feature maps at four scales: $\frac{1}{4}, \frac{1}{8}, \frac{1}{16}, \frac{1}{32}$. These feature maps are denoted as $F_s^1, F_s^2, F_s^3, F_s^4 $ and $F_t^1, F_t^2, F_t^3, F_t^4$.
Next, the lowest resolution feature maps $F_s^4$ and $F_t^4$ are passed through the global module G\_module to obtain the flow prediction $w^4$. Then, for each scale $\{(F_s^i,F_t^i)\}| i\in \{3,2,1\}$, the corresponding feature maps $F_s^i$ and $F_t^i$, along with the flow prediction $w^{i+1}$ from the previous layer, are input into the local module L\_module\_$i$. This allows us to obtain the flow prediction $w^i$ and change detection prediction $p^i$ for the current layer.
Finally, the flow prediction $w^1$ and change detection prediction $p^1$ are upsampled to $w^0$ and $p^0$ at the original resolution using bilinear interpolation.

The structures of the global module (G\_module) and the local modules (L\_modules) are depicted in Figure \ref{G_module and L_module}. More details are given below.

\begin{figure}[!htb]
    \centering
    \includegraphics[width=0.8\textwidth]{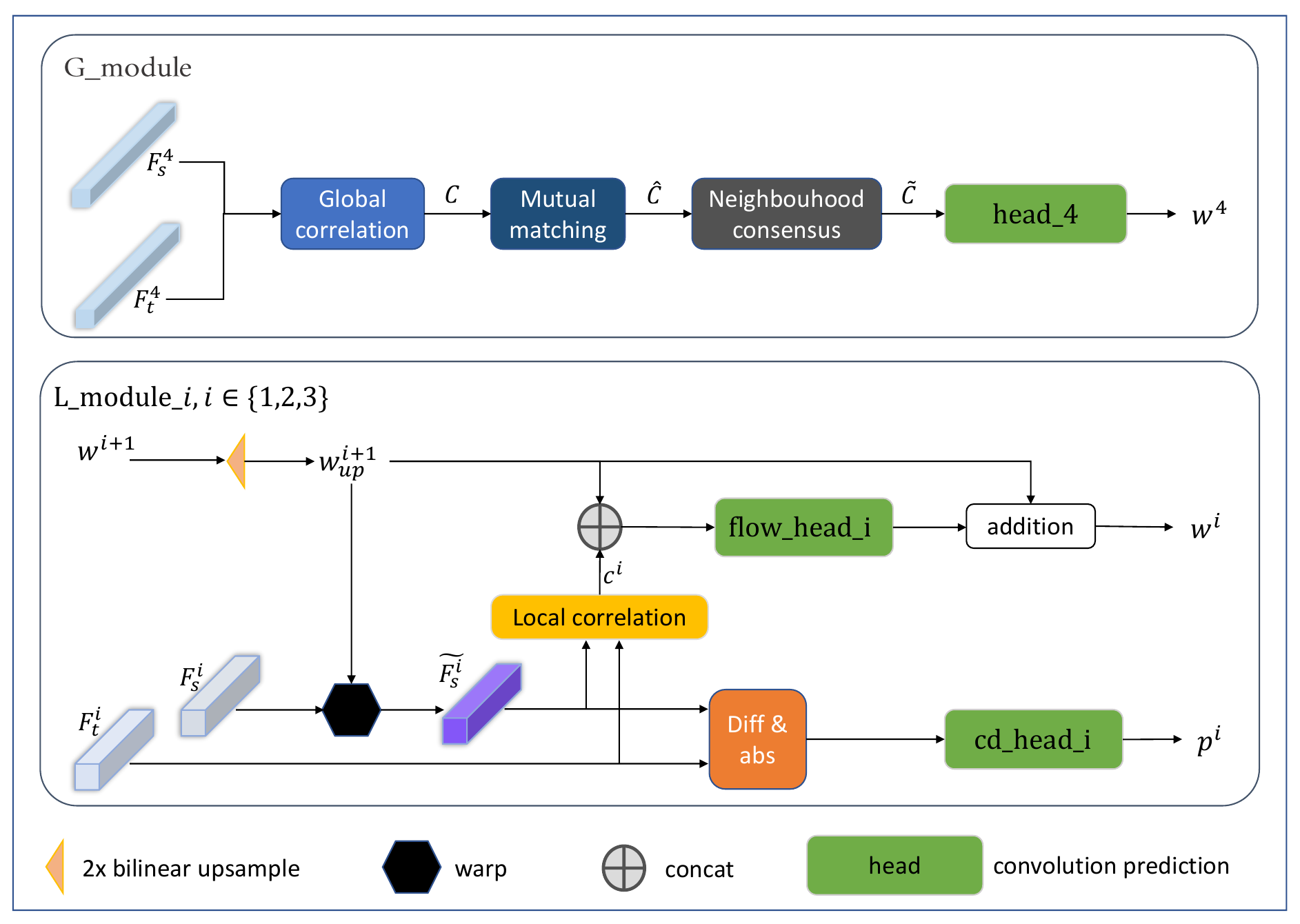}
    \caption{G\_module and L\_modules: G\_module predicts the flow field at scale $i=4$. L\_module takes $F_s^i$ and $F_t^i$ of the current layer, along with $w^{i+1}$ of the previous layer, as input and predicts the flow field $w^i$ and the change map $p^i$ of the current layer.}
    \label{G_module and L_module}
\end{figure}

\textbf{Global module G\_module:}
The G\_module comprises the global correlation layer, mutual matching layer, neighbourhood consensus layer, and prediction head head\_4. This module is specifically applied to the lowest resolution level $(F_s^4, F_t^4)$ due to its high computational requirements.

The global correlation layer is commonly employed to handle large displacements. It computes the similarity between the feature vectors of each position in $F_t^4$ and $F_s^4$, resulting in the global correlation map $C\in R^{\frac{H}{32}\times\frac{W}{32}\times\frac{H}{32}\times\frac{W}{32}}$. Mathematically, this can be expressed as:
\begin{equation}
C_{ijkl}=\frac{F_t^4 (i,j)^T F_s^4(k,l)}{\Vert F_t^4 (i,j)\Vert \Vert F_s^4(k,l) \Vert }.
\end{equation}

The mutual matching layer, proposed by \cite{9167479}, is utilized to mitigate false matches. It suppresses the similarity scores of incorrect matches and retains only the similarity scores of the best matches for each other. Mathematically, this can be expressed as:

\begin{align}
    \hat C_{ijkl} &=r_{ijkl}^s r_{ijkl}^t C_{ijkl} \\
    r_{ijkl}^s &= \frac{C_{ijkl}}{\max_{ab} C_{abkl}}\\
    r_{ijkl}^t &= \frac{C_{ijkl}}{\max_{cd} C_{ijcd}}.
\end{align}
 
The neighbourhood consensus layer is commonly employed to ensure that the prediction is independent of the order of the images in the image pair. Its calculation is as follows:
\begin{equation}
\tilde{C}=N(\hat{C})+N(\hat{C}^T),
\end{equation}
where $N$ is the neighborhood consensus network, which is a four-dimensional convolutional network. $\hat{C}^T_{ijkl}=\hat{C}_{klij}$.

Finally, a small convolutional network, head\_4, predicts the flow field $w^4$ as follows:  
\begin{equation}
w^4=\text{head\_4}(\tilde{C}).
\end{equation}

\textbf{Local module L\_module:} The L\_module for layer $i$ (where $i=1,2,3$) is denoted as L\_module\_$i$. L\_module\_1, L\_module\_2, and L\_module\_3 share the same structure but have different parameters. L\_module\_$i$ primarily consists of the upsampling operation, warp operation, local correlation layer, change detection predictor cd\_head\_$i$, and registration predictor flow\_head\_$i$. It takes $F_s^i$ and $F_t^i$ of the current layer, along with $w^{i+1}$ from the previous layer, as input and predicts the flow field $w^i$ and change map $p^i$ for the current layer. Below, we will provide a detailed description of the L\_module\_$i$ module.

Firstly, the flow field prediction $w^{i+1}$ from the previous layer is upsampled to $w_{up}^{i+1}$. This allows us to warp $F_s^i$ to $\tilde{F_s^i}$ using the following equation:
\begin{equation}
\tilde{F_s^i}(x)=F_s^i(x+w_{up}^{i+1}(x)),
\end{equation}
where $x$ is a two-dimensional position vector.

Next, L\_module\_$i$ branches into two paths to predict the flow field $w^i$ and the change detection probability map $p^i$, respectively.

For the flow field prediction branch, the local correlation layer is employed to model correspondences under small offsets:
\begin{equation}
c^i(x,d)=F_t^i(x)^T\tilde{F_s^i}(x+d),\max \vert d\vert \le r,
\end{equation}
where $x$ denotes the two-dimensional position coordinate, $d$ represents the position offset from $x$, and $r$ is the local search radius. In other words, the neighborhood corresponding to $x$ is a square area centered at $x$ with a side length of $2r+1$. The output $c^i \in R^{\frac{H}{2^{i+1}}\times\frac{W}{2^{i+1}}\times(2r+1)^2}$.

The flow field prediction $w^i$ is obtained using the residual mechanism:
\begin{equation}
w^i=w_{up}^{i+1}+\text{flow\_head\_}i(\text{Concat}(c^i,w_{up}^{i+1})),
\end{equation}
where flow\_head\_$i$ represents a small convolutional network.

For the change detection prediction branch, we calculate the absolute difference map between $\tilde{F_s^i}$ and $F_t^i$. This is followed by generating the change detection probability map $p^i$ using the cd\_head\_$i$ network. Mathematically, we have:

\begin{equation}
p^i=\text{cd\_head\_}i(\vert \tilde{F_s^i}-F_t^i\vert),
\end{equation}

where cd\_head\_$i$ represents a small convolutional network, and $\vert \cdot \vert$ denotes the absolute value operation applied to each element.

\section{Results}

\subsection{Evaluation Metrics}
In the end-to-end scenario, due to the possibility of inaccurate image registration, it is likely that pixels in the vicinity of a true positive pixel are predicted as belonging to the changed class. Therefore, we redefine the criteria for correctly predicting a positive and propose neighborhood-based change detection evaluation metrics.

For a pixel $a$ located at $(i,j)$ and belonging to the changed class, we define the square area with a radius of $r$ centered on $(i,j)$ as $D$. If at least one pixel in $D$ is predicted as belonging to the changed class, we consider $a$ to be correctly detected. Additionally, for all unchanged pixels in $D$, their corresponding values in the mask are set to 0, ensuring that they do not contribute to the evaluation. Thus, the neighborhood-based change detection evaluation metrics with respect to the radius $r$ are P@r, R@r, F1@r, IoU@r, and OA@r. When $r$ is set to 0, the above evaluation metrics are equivalent to those used in non-end-to-end change detection.

For image registration evaluation, we use the Percentage of Correct Keypoints (PCK-$\delta$) as the metric. Here, $\delta$ represents a percentage of the image size, and we set it to 0.05.

\subsection{Results: Two-stage Method vs E2ECDNet} 
The results obtained from the two-stage change detection method can be considered as an upper limit for the performance of the end-to-end change detection method. Therefore, we conducted experiments using both the two-stage and end-to-end approaches to compare their results. In the two-stage method, we first trained a single registration model on xBD-E2ECD and a single change detection model on registered xBD-E2ECD. These models were then cascaded for testing. To ensure a fair comparison, both the single registration model and single change detection model adopted the same architecture as E2ECDNet.

Table \ref{results of two-stage} presents the results of the two-stage method. It can be observed that when cascading the models to perform change detection on unregistered image pairs, the performance remains consistent compared to single change detection.

Table \ref{results on two-stage vs end-to-end} displays the results of the two-stage method compared to the end-to-end method (E2ECDNet). Our E2ECDNet achieves considerable R@5 but a lower precision rate compared to the two-stage method. This is because the registration model and change detection model in the two-stage method are trained individually and perform well on their respective tasks. However, in the end-to-end scenario, the presence of changed pixels increases the difficulty of image registration, which in turn affects the accuracy of change detection. Despite this, the two-stage method results in a more complex model and longer training time. On the other hand, our E2ECDNet can rapidly detect almost all changed areas and provide preliminary judgments for identifying damaged areas after natural disasters.

\begin{table}[!ht]
\caption{Results(\%) of the two-stage method on xBD-E2ECD.}
\label{results of two-stage}
\renewcommand\arraystretch{0.8}
\centering
\begin{tabular}{ c|l|l|l|l|l|l }
 \hline
two-stage & P & R & F1 & IoU & OA & PCK-0.05\\
 \hline
 single registration & -- & -- & -- & -- & -- & 79.01 \\
 single change detection & 34.40 & 79.36 & 47.99 & 31.57 & 88.42 & --\\
 cascade &34.37 & 79.40 & 47.97 & 31.56 & 88.41 & 79.01\\
 \hline
\end{tabular}
\end{table}

\begin{table}[!ht]
\caption{Results(\%) of the two-stage method vs the end-to-end the two-stage method vs the end-to-end method (E2ECDNet) on xBD-E2ECD.}
\label{results on two-stage vs end-to-end}
\renewcommand\arraystretch{0.8}
\centering
\begin{tabular}{ c|l|l|l|l|l|l|l}
 \hline
 method & r & P@r & R@r & F1@r & IoU@r & OA@r & PCK-0.05\\
 \hline
\multirow{2}{*}{two-stage} & 0 & 34.37 & 79.40 & 47.97 & 31.56 & 88.41 & 79.01\\
& 5 & 48.69 &  96.90 &  64.81 & 47.94 & 92.47 & 79.01 \\
\hline
\multirow{2}{*}{E2ECDNet} & 0 & 19.78 &  70.47 &  30.89  &  18.27  &  78.77  & 79.61 \\
& 5 & 29.22 &  98.25 &  45.04  &  29.07  &  82.83  & 79.61 \\
\hline
\end{tabular}
\end{table}

\begin{figure}[!ht]
     \centering
     \includegraphics[width=0.7\textwidth]{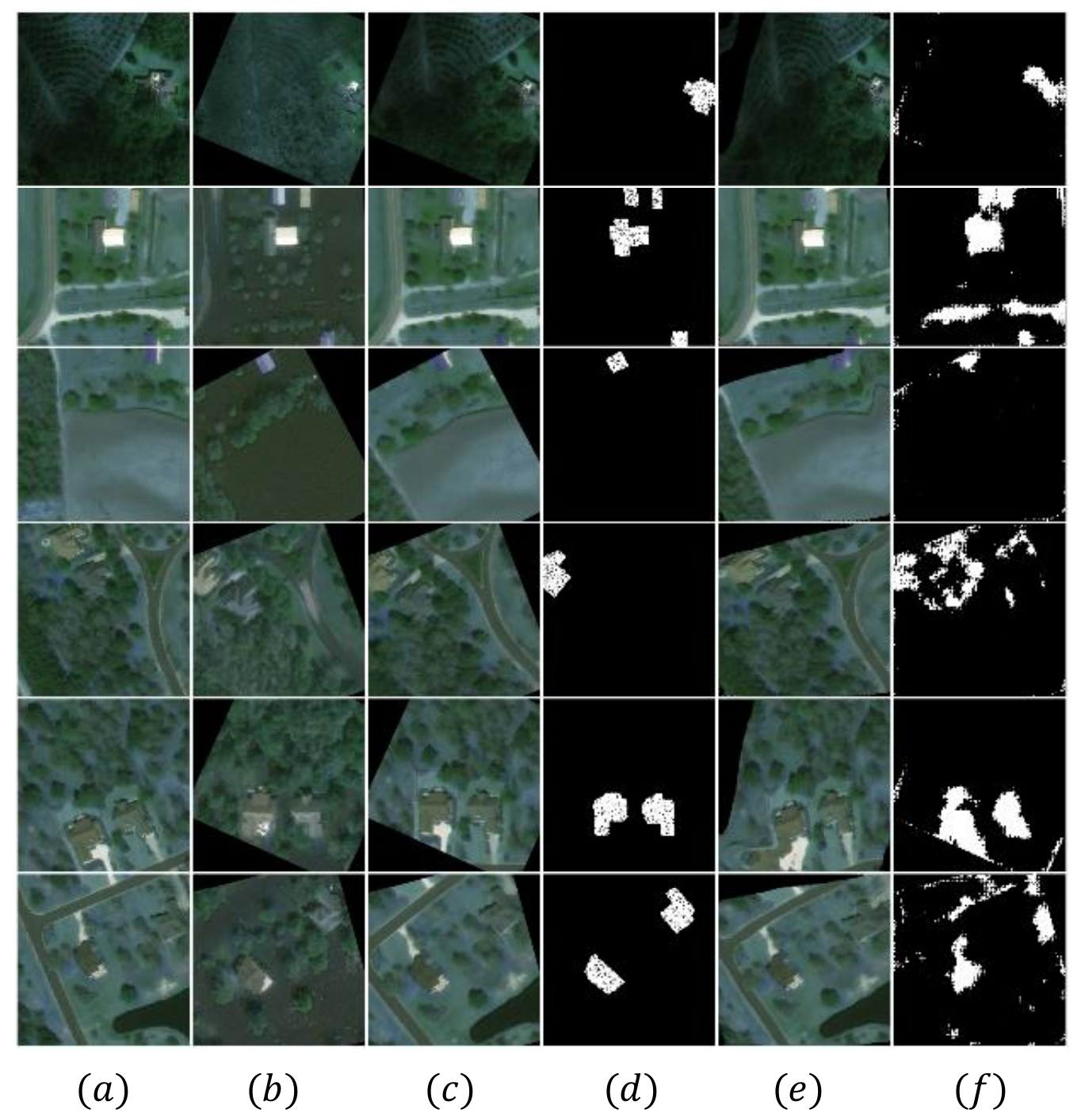}
     \caption{End-to-end change detection results of our E2ECDNet with resnet34 and multi-scale class balance cross entropy loss on the test set. (a) is the pre-event images, (b) is the post-event images, (c) is the images obtained by warping pre-event images according to the ground truth flow field, (d) is the ground truth change detection maps, (e) is the images obtained by warping the pre-event images according to the predicted flow field, (f) is the predicted change detection maps.}
     \label{result display 1}
 \end{figure} 
Figure \ref{result display 1} visually presents the end-to-end change detection results obtained by our E2ECDNet on the test set. Due to the inherent limitations of warping pre-event images to their corresponding positions in post-event images, not only are pixels belonging to the changed class predicted as changed, but also the surrounding pixels are often predicted as changed. Additionally, the inaccurate registration process makes pixels belonging to the unchanged class more susceptible to being misclassified as changed.

\section{Conclusions}
In this study, we have addressed the challenge of incomplete matching in bi-temporal image pairs for change detection in remote sensing. To overcome the limitations of existing registered datasets and the labor-intensive process of collecting unregistered image pairs with manual annotations, we have introduced a novel synthetic dataset called xBD-E2ECD, specifically designed for unregistered end-to-end change detection. 
Furthermore, we have proposed a novel end-to-end change detection network, E2ECDNet, which takes unregistered bi-temporal image pairs as input and simultaneously predicts the flow field and change map. To evaluate the performance of our approach, we have introduced a neighborhood-based evaluation standard that is well-suited for end-to-end change detection.
The experimental results have demonstrated the effectiveness of our proposed method. In our future work, we plan to investigate techniques to reduce the impact of pixel changes on registration and improve the precision of positive predictions.


\subsubsection{Acknowledgements} This work is supported by NSFC Key Projects of International (Regional) Cooperation and Exchanges under Grant 61860206004, and NSFC projects under Grant 61976201.
%
%
%
%




\bibliographystyle{splncs04}
\nocite{*}
\bibliography{reference}

\end{document}